\documentclass{guidelines}

\title{Annotation Guidelines for Corpus {\it Novelties}: Part 1 -- Named Entity Recognition}
\author{Arthur Amalvy \& Vincent Labatut}
\affiliation{Laboratoire Informatique d'Avignon -- LIA UPR 4128, Avignon, France}
\date{11 June 2024}
\version{1.0.1}

\abstractdef{The \textit{Novelties} corpus is a collection of novels (and parts of novels) annotated for Named Entity Recognition (NER). This document describes the guidelines applied during its annotation. It contains the instructions used by the annotators, as well as a number of examples retrieved from the annotated novels, and illustrating expressions that should be marked as entities as well as expressions that should \textit{not}.}

\begin{document}
\maketitle

%%%%%%%%%%%%%%%%%%%%%%%%%%%%%%%%%%%%%%%%%%%%%%%%%%%%
\newpage
\section{Introduction}
\label{sec:Intro}
This document aims at providing instructions for the annotation of named entities in the \textit{Novelties} corpus\footnote{\url{https://github.com/CompNet/Novelties}}. The corpus itself 
will be the object of a separate description. 
%is the object of a separate description~\cite{xxxxx}. 
It was constituted mainly to fulfill two goals: in the short term, train and test NER methods able to handle \textit{long} texts, and in the longer term, be used to develop \textit{Renard}~\citep{Amalvy2024a}, a pipeline aiming at extracting \textit{character networks} from literary fiction. This pipeline includes several processing steps after the NER, including coreference resolution and character unification. Character networks can be used to tackle a number of tasks, including the assessment of literary theories, the level of historicity of a narrative, detecting roles in stories, classifying novels, identify subplots, segment a storyline, summarize a story, design recommendation systems, align narratives, etc. See the detailed survey of \citet{Labatut2019} for more information regarding character networks.

This context drives the elaboration of the corpus, which explains why it exhibits certain differences with many similar NER corpora, such as \textit{CoNLL-2003}~\citep{Tjong2003} or \textit{OntoNotes v5}~\citep{Weischedel2011}. We originally based \textit{Novelties} on the literary corpus from~\citet{Dekker2019} as we describe in Section~\ref{sec:ApdxVersion} of the appendix. Note that there are other literary NER corpora (cf.~\cite{Ivanova2022} for a comparison), but they do not contain long texts~\cite{Dekker2019, Bamman2019} and/or do not fit our needs~\cite{Vala2016}. In addition, our end goal and the architecture of our pipeline affects the perimeter of what we consider to be a named entity, a point that we discuss further in Section~\ref{sec:IntroEntity}. Finally, these aspects also require us to put more emphasis on certain entity types, in particular characters, as explained in Section~\ref{sec:IntroTypes}. Our guidelines are based on similar instructions previously written for other annotation campaigns and corpora, both in French~\citep{Rosset2011, Soudani2018, Alrahabi2021} and English~\citep{Chinchor1998, LDC2008, Bamman2019}. We adapted them to fit our case and requirements.

%%%%%%%%%%%%%%%%%%%%%%%%%
\subsection{Notion of \textit{Named Entity}}
\label{sec:IntroEntity}
Historically, a named entity is a lexical unit of interest, which traditionally corresponds to a proper noun~\cite{Ehrmann2008}, and refers to an entity from the real world~\cite{Alrahabi2021}. Certain authors, such as \citet{Alrahabi2021} and~\citet{Bamman2019}, use more relaxed definitions, including (proper) definite descriptions~\cite{Ehrmann2008} in their annotations. A \textit{definite description} is an expression of the form \textit{determiner + noun phrase}, such as \textex{the President of the United States}. A \textit{proper} definite description allows identifying a \textit{unique} entity, e.g. \textex{the \textit{42nd} President of the United States}. \citet{Bamman2019} call them \textit{common entities}, by opposition to named entities.

On the one hand, we do not want to systematically annotate such expressions, because this quickly leads to \textit{nested} annotations. For instance, in the above example, \textex{United States} is contained in \textex{President of the United States}. Such nested entities, in turn, cause a number of technical complications. First, they make it much harder to define simple and consistent annotation rules: see the many rules and exceptions in the Quaero guidelines~\cite{Rosset2011}, for instance. Second, they do not allow the traditional tags-based representation on which many models are based. But on the other hand, in certain novels, some major characters are exclusively referred to through definite expressions. Of course, we do not want to miss any important character, even if it is not properly named in the story. For this reason, there are some exceptions for which we annotate definite expressions in addition to proper nouns.

Even with these clarifications, it is not always obvious to determine what is a named entity and what is not. As stated by~\citet{Mcdonald1993}, one can distinguish two types of evidence that are helpful in order to come to a decision regarding the annotation of an entity. \textit{Internal} evidence is directly present in the expression of interest. It can consist of criteria such as capitalization, the inclusion of a known name or the presence of titles or abbreviations for corporation types such as \textit{``Ltd.''}. By comparison, \textit{external} evidence is found in the context surrounding the expression of interest: in a novel, the description of a character's actions is evidence that some of their mentions are named entities. We conducted a first annotation pass over a few chapters to get a beta version of our corpus (cf. Appendix~\ref{sec:ApdxVersion}), and leveraged this experience to identify four types of evidence (two internal and two external) that help to make this decision in the context of novels. While internal evidence is easily interpreted locally, the interpretation of external evidence may necessitate access to knowledge from the entire novel or from its universe, something that can often be done through the help of online wikis dedicated to specific literary universes. Most of the four types of evidence we describe below are neither necessary nor sufficient to really detect an entity, but rather serve as important hints for annotators.

%%%
\paragraph{Capitalization (internal evidence)} 
The expression is (possibly partly) capitalized. In English and French, proper nouns are capitalized, so this is a good indication that the expression is a named entity. Some authors are very liberal in their use of capitals, though, so an upper-cased initial does not necessarily mean that the expression is a named entity.

%%%
\paragraph{Self-Sufficiency (internal evidence)} 
The expression alone has a meaning and refers to an entity or a group of entities. Contrary to the other three factors, this one is necessary for an entity to be considered as valid. This point helps when dealing with parts of proper nouns, e.g. first names.

%%%
\paragraph{Unicity (external evidence)} 
The expression serves to uniquely identify an entity (or possibly a group of entities). This point is related to the notion of \textit{proper} definite expression. The goal here is to exclude generic expressions such as \textex{a policeman}, \textex{a little girl}, etc. Unicity is closely linked to the fact that we consider the universe of a literary text as a \textit{closed world}, where two distinct entities would be clearly identified differently, and may not apply to other kinds of texts.

%%%
\paragraph{Frequency (external evidence)} 
The expression is frequently used to refer to the entity. It is not possible to define an absolute threshold above which an expression should be considered as frequent enough, so this is left to the appreciation of the annotator. The point is to catch expressions such as nicknames, that are not proper nouns, but still used to refer to certain characters.

%%%%%%%%%%%%%%%%%%%%%%%%%
\subsection{Considered Entity Types}
\label{sec:IntroTypes}
Detecting a named entity in the text is only the first part of the task: the second part consists in determining its category or type. Historically, NER authors are interested in distinguishing between persons, locations, and organizations. Moreover, in many works, the expression \textit{named entity} refers not only to proper nouns, but also to temporal expressions (dates and times) and quantities (amounts of money, percentages)~\cite{Ehrmann2008}.

In our case, because of our end goals, we have a strong focus on one specific category of named entities: those referring to characters. However, we decided to include other types of entities in our annotation, too, as this task does not require much additional work, while increasing the value of the corpus.

%%%
\paragraph{Characters} 
This type supersedes the traditional \textit{Persons} category: it includes the usual \textit{anthroponyms} (names of people), but also other entities referring to other types of agents, possibly non-human, such as animals, robots, magical creatures, etc.

%%%
\paragraph{Locations} 
This is the standard category used in many corpora, which includes \textit{toponyms} (i.e. names of places). We do not distinguish between natural (river, mountain, island...) and artificial (city, street, building, etc.) locations.

%%%
\paragraph{Organizations}
This type is also standard, in the sense that it appears in many corpora. It gathers all named entities referring to explicitly organized institutions, such as a government, a company, etc. It excludes informal groups such as families, \textit{ethnonyms} (names given to the members of ethnic groups), and \textit{demonyms} (names given to the inhabitants of some places).

%%%
\paragraph{Groups}
This type is more uncommon: it gathers informal groups of persons referred to under the same umbrella name, including family names, ethnonyms, and demonyms. It is used when the text does not identify its members individually. Traditionally, NER corpora annotate these kinds of expressions as \textit{Persons}, but we differentiate them in order to facilitate character detection.

%%%
\paragraph{Miscellaneous}
This heterogeneous category gathers \textit{praxonyms} (names referring to historical, cultural, commercial or sport events), \textit{ergonyms} (names of objects and man-made products) provided they are unique, \textit{phenonyms} (names of meteorological events such as tempests), and titles of intellectual or cultural works (such as books and movies).

%%%%%%%%%%%%%%%%%%%%%%%%%
\subsection{Document Conventions}
\label{sec:IntroConv}
In the rest of the document, we provide a number of examples to illustrate our guidelines. Offline examples are represented in gray frames, as follows:
\begin{example}
    This is an example.
\end{example}
\noindent Inline examples are inserted in the text using a sans-serif font, e.g. \textex{the Emperor}.

We use boxes to highlight entities in the text. Their color indicates the entity type, and we increase the transparency to distinguish between entities that are the focus of an example, and those that just happen to appear in the example :
\vspace{-0.25cm}
\begin{center}
    \begin{tabular}{l l l l l}
        \textit{Characters} & \texttt{CHR} & Red & \entCHR{Elric} & \entCHRwm{Elric} \\[1mm]
        \textit{Locations} & \texttt{LOC} & Blue & \entLOC{Avignon} & \entLOCwm{Avignon} \\[1mm]
        \textit{Organizations} & \texttt{ORG} & Brown & \entORG{Mozilla Foundation} & \entORGwm{Mozilla Foundation} \\[1mm]
        \textit{Groups} & \texttt{GRP} & Orange & \entGRP{Harkonnens} & \entGRPwm{Harkonnens} \\[1mm]
        \textit{Other entities} & \texttt{MSC} & Green & \entMSC{Emacs} & \entMSCwm{Emacs} \\
    \end{tabular}
\end{center}
\vspace{-0.20cm}\noindent When we want to specifically highlight that an expression should \textit{not} be annotated, we use a gray box, e.g. \entNOT{the boy}.

%%%%%%%%%%%%%%%%%%%%%%%%%
\subsection{Organization}
\label{sec:IntroOrg}
In the following, we first describe guidelines that are applicable for all the types of entities that we annotate (Section~\ref{sec:Gral}). Afterward, each remaining section is dedicated to a specific type of entity: characters (Section~\ref{sec:Chr}), locations (Section~\ref{sec:Loc}), organizations (Section~\ref{sec:Org}), groups (Section~\ref{sec:Grp}), and other entities (Section~\ref{sec:Msc}). In Section~\ref{sec:Confus}, we discuss the possible confusion between certain types of entities. Finally, Section~\ref{sec:Conclu} provides our concluding remarks, and Appendix~\ref{sec:ApdxVersion} gives the history of this document.

%%%%%%%%%%%%%%%%%%%%%%%%%%%%%%%%%%%%%%%%%%%%%%%%%%%%
\newpage
\section{General Principles}
\label{sec:Gral}
This section describes some general rules that apply to all entities, independently of their type. There are some exceptions to these rules, which are described latter, in type-specific sections.

Annotation is conducted manually, and the human aspect of this process must therefore be taken into account, in addition to the more technical points mentioned before. In particular, the annotation rules must be clear enough, simple enough, and not too numerous, in order to avoid human errors. For this reason, we sometimes sacrifice accuracy if it allows providing the annotator with simpler or more consistent instructions. We also noticed that human readers tend to \textit{want} to annotate certain expressions. Not providing any instructions for these cases, or forbidding to annotate them, can be counterproductive. For instance, in an earlier version of these guidelines, we did not annotate languages at all. But annotators did it anyway, probably because the same word is generally used to refer to a language and its speakers (and the latter were already annotated as a group). Therefore, in a subsequent version of the guidelines, we included language annotation as a miscellaneous entity.

%%%%%%%%%%%%%%%%%%%%%%%%%
\subsection{Nested Entities}
\label{sec:GralNested}
Nested entities are entities within entities, e.g. \textex{President of the United States}: the \textex{United States} part is an organization, but the whole expression is a character. As explained in Section~\ref{sec:Intro}, detecting \textit{nested} entities is quite different from standard NER, almost a different problem. For this reason, as in many guidelines~\cite{Finkel2009}, we focus only on \textit{flat} entities in this corpus. This means making a choice between the different levels of entities within the nested structure. 

%%%
\paragraph{General Rule}
Certain authors keep the innermost entity, e.g.~\cite{Alrahabi2021}, whereas other focus on the outmost, as explained in~\cite{Finkel2009}. It appears that, in novels, the outmost entity is generally the entity participating in the action, or the entity that the author wants to mention. By comparison, the innermost entity brings some additional information allowing to identify the outmost entity, but is not the main object of the mention. For this reason, our general rule is to annotate the outmost entity.

%%%
\paragraph{Examples}
See this case from Aldous Huxley's \textit{Brave New World}, for instance:
\begin{example}
    the \entCHR{Director of Hatcheries and Conditioning} entered the room, in the scarcely breathing silence %\textelp{}
\end{example}
\noindent This is one of the main characters in the novel. He is also called by his first name (\textex{Thomas}) and nickname (\textex{Tomakin}), but more importantly by the acronym \textex{D.H.C.}, which shows the importance of including the organization \textex{Hatcheries and Conditioning} in the annotation. Annotating the innermost entity, i.e. only this organization, would be much less informative. 

Interestingly, his full title is actually \textex{Director of Hatcheries and Conditioning of Central London}, but it is never used under this form. We rather find:
\begin{example}
    The \entCHR{D.H.C. for Central London} always made a point of personally conducting his new students %\textelp{}
\end{example}
\noindent Although there is no other \textex{D.H.C.} in the novel, and therefore no possible confusion, we annotate \textex{Central London} as a part of the entity, for the sake of consistency.

Nestedness also concerns other types of entities than characters. Regarding organizations, we can mention the \textex{Knight of the Vale}, from \textit{A Song of Ice and Fire}, the Fantasy series by George R. R. Martin:
\begin{example}
    ``The \entORG{Knights of the Vale} could make all the difference in this war,'' said \entCHRwm{Robb} \textelp{}
\end{example}
\noindent Where the \textex{Vale} is a location. The full form of this name is actually the \textex{Vale of Arryn}, where \textex{Arryn} refers to a person, making this a nested entity, too:
\begin{example}
    \textelp{} there were no friends of the \entGRPwm{Lannisters} in the \entLOC{Vale of Arryn}.
\end{example}

%%%
\paragraph{Counterexamples}
There are some exceptions to this general rule, though. When the outmost entity is deemed too generic and/or not frequent, we annotate the outmost \textit{valid} entity. In practice, there are often only two levels of nestedness, so this amounts to selecting the innermost entity. See this example from Glen Cook's \textit{The Black Company}:
\begin{example}
    \textelp{} \entCHRwm{Bucket}'s answer. ``He wanted to pick off \entNOT{Black Company guys}. That's obvious.''
\end{example}
\noindent Here, one could want to annotate \textex{Black Company guys} as a group. However, if the expression may be frequent, it is also very imprecise. Over the series, it is likely to refer to several distinct subsets of the people constituting the \textex{Black Company}. For this reason, in this case, we would annotate only \textex{Black Company}, as a an organization.

In another example, this time from \textit{A Song of Ice and Fire}, the situation is different :
\begin{example}
    ``\entCHRwm{Daenerys Targaryen} has wed some \entNOT{Dothraki horselord}. \textelp{} Shall we send her a wedding gift?''
\end{example}
\noindent Here, \textex{Dothraki horselord} refers to the character \textex{Khal Drogo}. It is precise enough, but used only a few times over the whole series, so not considered as frequent. Consequently, we only annotate \textex{Dothraki}, as a group (see Section~\ref{sec:GrpDemonyms}).

The same remarks apply for other entity types. See this example from \textit{The Black Company}, that involves a location:
\begin{example}
    \textelp{} the avenue \textelp{} winds from the \entLOCwm{Customs House} uptown to the \entNOT{Bastion's main gate}.
\end{example}
\noindent The expression \textex{Bastion's main gate} could be annotated as a location, because this name is quite precise. However, it is not frequent at all. We consider it more informative to annotate only \textex{Bastion} as a location.

Similarly, in the following example from J. K. Rowling's \textit{Harry Potter}:
\begin{example}
    \textelp{} and a moment later, \entNOT{Dudley's best friend}, \entCHRwm{Piers Polkiss}, walked in with his mother.
\end{example}
\noindent Here, \textex{Dudley's best friend} refers to a specific entity, which is precisely identified. However, this expression is not frequent (used only once), which is why we annotate \textex{Dudley} as a character. We generally do not annotate expressions that refer to an entity through its relation to another entity (here: \textex{Piers} through \textex{Dudley}), except when it is very frequent.

%%%
\paragraph{Enumerations}
Nested entities should be distinguished from \textit{enumerations} of entities that share a part of their name. In this case, we annotate each entity separately, provided the expression is sufficient to recognize it:
\begin{example}
    \textelp{} with both hands and said, ``In the name of \entCHR{Robert of the House Baratheon}, the First of his Name, \entCHR{King of the Andals and the Rhoynar and the First Men}, \entCHR{Lord of the Seven Kingdoms} and \entCHR{Protector of the Realm}, by the word of \entCHR{Eddard of the House Stark},\entCHR{ Lord of Winterfell} and \entCHR{Warden of the North}, I do sentence you to die.''
\end{example}

Otherwise, depending on the case, some parts may not be annotated, or the whole expression may be considered as a group (see Section~\ref{sec:Grp}).

%%%%%%%%%%%%%%%%%%%%%%%%%
\subsection{Definite Descriptions}
\label{sec:GralDescr}

%%%
\paragraph{General Rule}
As stated in the introduction, we generally do not annotate definite descriptions. This purposely excludes generic mentions, such as \textex{the boy} or \textex{the city} in the following excerpts of Brandon Sanderson's \textit{Elantris}:
\begin{example}
    \begin{itemize}[leftmargin=*,itemsep=1mm]
        \item \entNOT{The boy}, as if realizing his chance would soon pass, stretched his arm forward \textelp{}
        \item He had hoped \entNOT{the city} would grow less gruesome as he left the \entNOT{main courtyard} \textelp{}
    \end{itemize}
\end{example}

There are several exceptions to this general rule, though, which we detail in the entity-specific sections. In principle, if a definite expression is frequently used to refer to a specific entity, then it can be annotated. For instance, for a character, this expression could be considered as a nickname (see Section \ref{sec:ChrNick}).

%%%
\paragraph{Capitalization}
As discussed in Section~\ref{sec:IntroEntity}, capitalization is a good indication that a definite expression should be annotated as an entity. For instance, in the below sentence from the Fantasy series \textit{The Black Company}, the capitals hint at a \texttt{LOC} entity, and not just any undifferentiated hill:
\begin{example}
    ``We're going to the \entLOC{Necropolitan Hill} to eyeball that \entGRPwm{forvalaka} tomb.''
\end{example}
\noindent However, this principle does not always applies, as the use of capitals vary widely from one author to the other. For instance, in Aldous Huxley's \textit{Brave New World}:
\begin{example}
    \textelp{} until at last they were dancing in the crimson twilight of an \entNOT{Embryo Store} \textelp{} 
\end{example}
\noindent The use of the indefinite article \textex{an} clearly indicates that, despite the capitalization, the author mentions an arbitrary \textex{embryo store}, and not a specific, recurrent place. The frequency rule mentioned above helps deciding whether that expression is recurring or not.

%%%%%%%%%%%%%%%%%%%%%%%%%
\subsection{Determiners}
\label{sec:GralDet}

%%%
\paragraph{General Rule}
Except when they are explicitly part of the name they are attached to, we do not annotate determiners in front of entities. This is because some of these entities can be referred to without determiners. 

%%%
\paragraph{Examples}
In \textit{The Black Company} series, the eponymous organization is referred too as \textex{the Black Company} in the novels, but also sometimes as only \textex{Black Company}, depending on context. This shows that the determiner is not crucial to the designation of this entity. Therefore, we keep its smallest consistent expression:
\begin{example}
    The \entORG{Black Company} does not suffer malicious attacks upon its men.
\end{example}
\noindent In the same series, the \textex{Lady} is one of the main characters. The same rule applies:
\begin{example}
    \entORGwm{Oar} had not yet seen any of the \entCHR{Lady}'s champions.
\end{example}

%%%
\paragraph{Counterexample}
On the contrary, sometimes the determiner is part of the name, in which case we include it in the annotation to be consistent the rule of self-sufficiency:
\begin{example}
    He took the train to \entLOC{The Hague}.
\end{example}

%%%%%%%%%%%%%%%%%%%%%%%%%
\subsection{Parts of Names}
\label{sec:GralParts}

%%%
\paragraph{General Rule}
Sometimes, an entity is mentioned through a part of its name, instead of its full name. In this case, we annotate this part, but only under the condition that this incomplete name is sufficient to identify the entity, and that it is used frequently. 

%%%
\paragraph{Examples}
This particularly apply to characters, when using only a first name, e.g. in the \textit{Harry Potter} series, the eponymous character is often called only by his first name only:
\begin{example}
    ``I've come to bring \entCHR{Harry} to his aunt and uncle. They're the only family he has left now.''
\end{example}

But the situation also happens for other types of entities, in particular organizations. See this example from \textit{The Black Company}, where the eponymous organization is only called by an abbreviated (but unambiguous) version of its full name:
\begin{example}
    He and \entCHRwm{One-Eye} have been with the \entORG{Company} a long time.
\end{example}

%%%%%%%%%%%%%%%%%%%%%%%%%
\subsection{Misspelled Names}
\label{sec:GralMisc}

%%%
\paragraph{General Rule}
There are several situations where the name of an entity is not correctly spelled, in which our general rule is to annotate the mention as if it was correctly written.

%%%
\paragraph{Examples}
In Joe Abercrombie's \textit{The Blade Itself}, one of the character has a speech impediment, and some character names are written in a way that reflect this trait:
\begin{example}
    `Ith \entCHR{Theverar},' \textelp{} by which \entCHRwm{Glokta} understood that \entCHRwm{Severard} was at the door.
\end{example}
\noindent Here, the name of character \textex{Severard} is rendered as \textex{Theverar}. Same thing with \textex{Felix Grandet}'s fake stuttering in Balzac's \textit{Eugénie Grandet}:
\begin{example}
    ``\entCHR{M-m-monsieur de B-B-Bonfons},'' --for the second time in three years \entCHRwm{Grandet} called \textelp{} %the \entCHRwm{Cruchot} nephew \textelp{}
\end{example}

In the following example from Herman Melville's \textit{Moby Dick}, the non-standard spelling is rather a matter of accent:
\begin{example}
     ``Passed one once in \entLOC{Cape-Down},'' said the old man sullenly.
\end{example}
\noindent The speaker is \textex{Fleece}, the cook of the ship, and by \textex{Cape-Down} he means \textex{Cape Town}.

Sometimes, the misspelling of a name can be due to the speaker's error. Here is an example from \textit{Moby Dick}, in which \textex{Captain Peleg} makes a mistake when saying the name of a character:
\begin{example}
    I say, \entCHR{Quohog}, or whatever your name is, did you ever stand in the head of a whale-boat?
\end{example}
\noindent The character's actual name is \textex{Queequeg}, but we annotate the wrong name like before, as it is obvious who \textex{Peleg} talks to, given the context.

%%%%%%%%%%%%%%%%%%%%%%%%%%%%%%%%%%%%%%%%%%%%%%%%%%%%
\newpage
\section{Character Entities (\texttt{CHR})}
\label{sec:Chr}
The standard approach to annotate characters would be to consider them as persons, and to use the very common \texttt{PER} tag. However, as remarked by \citet{Bamman2019} when annotating LitBank, characters are not necessarily persons. For this reason, they use a wider definition and annotate all entities who ``engage in dialogue or have reported internal monologue, regardless of their human status''. They still consider them formally as persons, though, and use \texttt{PER}. 

Many works of fiction involve non-human agents that have an effect on the story. Therefore, we go further, and annotate any individual entity with some form of sentience and agency in the plot. As a consequence, contrary to other classical NER datasets, we do not annotate \textit{persons}, but rather \textit{characters}. This wider concept encompasses not only human entities, but also other sentient entities such as animals, mythical creatures, magical weapons, robots\ldots{} To stress this difference, we use a specific tag, \texttt{CHR}, instead of the traditional \texttt{PER}.

%%%%%%%%%%%%%%%%%%%%%%%%%
\subsection{Proper Nouns}
\label{sec:ChrProper}

%%%
\paragraph{General Rule}
We annotate proper nouns that refer to individual characters, e.g. in Jane Austen's \textit{Emma}:
\begin{example}
    \entCHR{Emma Woodhouse}, handsome, clever and rich, with a comfortable home \textelp{}
\end{example}

%%%
\paragraph{Parts of Names}
As explained in Section~\ref{sec:GralParts}, it is possible to annotate isolated parts of the name, provided they allow identifying the character without ambiguity. For example, also from Jane Austen's \textit{Emma}:
\begin{example}
    \textelp{} and \entCHR{Emma} could not but sigh over it, and wish for impossible things \textelp{}
\end{example}

In Dostoevsky's \textit{The Double}, the main character \textex{Yakov Petrovich Golyadkin} is mentioned by various combinations of parts of his name:
\begin{example}
    \begin{itemize}[leftmargin=*,itemsep=1mm]
        \item A man with a message. ``Is \entCHR{Yakov Petrovitch Golyadkin} here?'' says he.
        \item ``He's still at the office and asking for you, \entCHR{Yakov Petrovitch}.''
        \item ``You're mischievous \entCHR{brother Yakov}, you are mischievous!''
        \item When he had made this important discovery \entCHR{Mr. Golyadkin} nervously closed his eyes \textelp{}
    \end{itemize}
\end{example}

%%%
\paragraph{Antonomasia}
Certain authors mention person names through \textit{antonomasia}, a metonymy consisting in using a proper noun as a common name. It is questionable whether the mentioned person should be considered as a proper character, or just a cultural reference. We decide to annotate such cases when the author uses an initial capital.

Here is an example from \textit{Moby Dick}:
\begin{example}
    I laugh and hoot at ye, ye cricket-players, ye pugilists, ye deaf \entCHR{Burkes} and blinded \entCHR{Bendigoes}!
\end{example}
\noindent where \textex{Burket} and \textex{Bendigo} are 19\textsuperscript{th} century boxers.

%%%
\paragraph{Groups}
If a name refers to several characters at once, we annotate the entity as a \textit{group} instead (see Section~\ref{sec:Grp}. Consider, for instance, this excerpt of \textit{Harry Potter}:
\begin{example}
    They didn't think they could bear it if anyone found out about the \entGRP{Potters}.
\end{example}
\noindent Here, \textex{the Potters} collectively refers to \textex{James Potter}, \textex{Lily Potter} and \textex{Harry Potter}.

%%%%%%%%%%%%%%%%%%%%%%%%%
\subsection{Presence vs. Evocation}
\label{sec:GralPres}
Generally speaking, it is possible to explicitly annotate whether a character is present or evoked, as in certain guidelines like~\cite{Alrahabi2021}. In the former case, the character is physically present and participating in the scene, like in this example from Aldous Huxley's \textit{Brave New World}:
\begin{example}
    \entCHR{John} began to understand. ``Eternity was in our lips and eyes,'' he murmured.
\end{example}
\noindent In the latter case, the character is just brought up by other entities in their absence, as in this example from the same novel:
\begin{example}
    ``I suppose \entCHR{John} told you. What I had to suffer --and not a gramme of \entMSCwm{soma} to be had.
\end{example}

%%%
\paragraph{General Rule}
In the context of these guidelines, we assume that distinguishing both types of entity mentions (presence vs. evocation) can be done in a later step of our pipeline mentioned in Section~\ref{sec:Intro}. Consequently, as a general rule, we annotate indifferently both situations.

%%%
\paragraph{Interjections}
However, it happens that the name of a person is used as an interjection. This is particularly the case of divinities, e.g. in Dostoevsky's \textit{The Double}:
\begin{example}
    ``I'm very well, thank \entCHR{God}, \entCHRwm{Anton Antonovitch},'' said \entCHRwm{Mr. Golyadkin}, stammering.
\end{example}
\noindent Ideally, it would make sense to ignore such mentions, as \textex{God} is not a character actually participating in the story, in this case. However, this decision could be considered too subjective. Therefore, to simplify the annotation task, we decide to annotate all these invocations as characters too. A specific step of our pipeline could determine later whether one entity should be kept, depending on it being a proper character.

The novel \textit{Brave New Workd} exhibits an interesting case of divine invocation, as \textex{Henri Ford}'s name is almost always used in place of \textex{God}'s, as an interjection:
\begin{example}
     ``Oh, \entCHR{Ford}!'' he said in another tone, ``I've gone and woken the children.''
\end{example}
\noindent As explained before, we annotate \textex{Ford} as a character even if he does not intervene directly in the story.

%%%
\paragraph{Special Case}
The distinction is sometimes more fuzzy, as certain novels involve divinities as characters while also using their names in interjections. This is the case of \textex{God} in Douglas Adams's \textit{Hitchhiker's Guide to the Galaxy}:
\begin{example}
    \begin{itemize}[leftmargin=*,itemsep=1mm]
        \item \entCHR{God}, what a terrible hangover it had earned him though.
        \item ``Oh dear,'' says \entCHR{God}, ``I hadn't thought of that,'' and promptly vanishes in a puff of logic.
    \end{itemize}
\end{example}
\noindent Another example is \textex{Hood}, the god of death in Steven Erikson's \textit{Malazan Book of the Fallen} Fantasy series:
\begin{example}
    \begin{itemize}[leftmargin=*,itemsep=1mm]
        \item Clear the streets? How in \entCHR{Hood}'s name do we manage that?
        \item \entCHR{Hood} glanced down at the spatter on its frayed robes.
    \end{itemize}
\end{example}
\noindent It is difficult for the reader to guess whether these divinities are supernaturally permanently listening to the people, and hear them pronouncing their names. For the sake of simplicity, we annotate not only situations where the divinity appears explicitly as a character, but also interjections, as in the above examples.

%%%%%%%%%%%%%%%%%%%%%%%%%
\subsection{Definite Descriptions}
\label{sec:ChrDescr}

%%%
\paragraph{General Rule}
As explained more generally in Section~\ref{sec:GralDescr}, contrary to \citet{Bamman2019}, we do not annotate definite descriptions, in general:
\begin{example}
    He was still determined to not mention anything to \entNOT{his wife}.
\end{example}

%%%
\paragraph{Exceptions}
Some characters are only mentioned using a definite description. For instance, this is particularly true for Carlo Collodi's \textit{The Adventures of Pinocchio}, in which many characters are never properly named: \textex{the Judge}, \textex{the Innkeeper}, \textex{the Falcon}, \textex{the Owl}, \textex{the Farmer}, etc. In these cases, we annotate such expressions:
\begin{example}
    The \entCHR{Judge} was a Monkey, a large Gorilla \textelp{} The \entCHR{Judge} listened to him with great patience.
\end{example}

Another example is the already previously discussed \textex{Director of Hatcheries and Conditioning} in \textit{Brave New World}, which is very often called just \textex{Director}. There are just two other directors in the whole novel, and each one is mentioned only once. For this reason, we annotate \textex{Director} as a character, as there is no ambiguity, and the use is frequent:
\begin{example}
    Tall and rather thin but upright, the \entCHR{Director} advanced into the room.
\end{example}
\noindent This case is also related to the situation where we annotate societal roles (cf. Section~\ref{sec:ChrRoles}).

%%%%%%%%%%%%%%%%%%%%%%%%%
\subsection{Titles \& Honorifics}
\label{sec:ChrTitles}

%%%
\paragraph{General Rule}
We annotate honorific titles as part of \texttt{CHR} entities, even when they are lowercase. This choice is driven by our end application (character network extraction), where titles carry important information: they can be used to disambiguate between several characters, or to detect their gender.

%%%
\paragraph{Examples}
Consider the following sentences, for which the general rule directly applies:
\begin{example}
    \begin{itemize}[leftmargin=*,itemsep=1mm]
        \item We talked it all over with \entCHR{Mr. Weston} last night.
        \item \entCHR{Lord Eddard Stark} dismounted and his ward \entCHRwm{Theon Greyjoy} brought forth the sword.
    \end{itemize}
\end{example}
\noindent The first sentence comes from \textit{Emma}, and the second from \textit{A Song of Ice and Fire}.

Titles are sometimes necessary to distinguish between certain characters. For instance, in Balzac's \textit{Eugénie Grandet}:
\begin{example}
    Monsieur and \entCHR{Madame Guillaume Grandet}, by gratifying every fancy of their son \textelp{}
\end{example}
\noindent Here, \textex{Monsieur} is not annotated because it cannot stand by itself (see the \textit{Isolated Titles} paragraph, below). \textex{Madame Guillaume Grandet} refers to \textex{Guillaume Grandet}'s wife, and without the title \textex{Madame}, this mention would be mistakenly understood as referring to her husband.

During the elaboration of these guidelines, we considered annotating titles separately from characters, as a distinct entity type. However, this would be very close to handling nested entities, which we want to avoid (see Section~\ref{sec:GralNested}).

Titles include family-related relations. For instance, in \textit{A Song of Ice and Fire}, \textex{Jon Snow} is the nephew of \textex{Benjen Stark}:
\begin{example}
    \entCHR{Uncle Benjen} studied his face carefully. ``The \entLOCwm{Wall} is a hard place for a boy, \entCHRwm{Jon}.''
\end{example}

Honorific titles can be completely fictional, like for instance \textex{High Fist} in the \textit{Malazan Book of the Fallen} Fantasy series:
\begin{example}
    \entCHR{High Fist Dujek Onearm} entered, the soap of his morning shave still clotting the hair in his ears.
\end{example}

%%%
\paragraph{Isolated Titles}
Since entities must be self-sufficient, we do not annotate isolated titles as \texttt{CHR}, in general:
\begin{example}
Thank you, \entNOT{sir}! Please, come again.
\end{example}
\noindent We make an exception: it is possible to consider such an isolated title as unique and frequent, similarly to what we do with nicknames in Section~\ref{sec:ChrNick}. This case is very close to the annotation of societal roles that we describe in Section~\ref{sec:ChrRoles}.

%%%%%%%%%%%%%%%%%%%%%%%%%
\subsection{Nickames}
\label{sec:ChrNick}

%%%
\paragraph{General Rule}
We annotate nicknames if they are frequent and allow identifying the entity in a reasonably unique way (see Section~\ref{sec:IntroEntity}).

%%%
\paragraph{Examples}
For instance, in \textit{A Song of Ice and Fire}, character \textex{Mance Rayder} is often referred to as follows:
\begin{example}
    \textelp{} he was a wildling, his sword sworn to \entCHRwm{Mance Rayder}, the \entCHR{King-Beyond-the-Wall}.
\end{example}
\noindent Another example is \textex{White Whale} (although this could also be considered a definite decription), an expression frequently used to refer to the eponymous whale in \textit{Moby Dick} (note the capitalization):
\begin{example}
    \textelp{} many brave hunters, to whom the story of the \entCHR{White Whale} had eventually come.
\end{example}
\noindent It is possible that the nickname concerns only a part of the original name, e.g. the first name for \textex{Eddard Stark}:
\begin{example}
    ``You're \entCHR{Ned Stark}'s bastard, aren't you?'' \entCHRwm{Jon} felt a coldness pass right through him.
\end{example}
\noindent Incidentally, observe that we do not annotate the whole expression \textex{Ned Stark's bastard} as a character, because it is not frequent enough (see Section~\ref{sec:GralNested}).

Some characters have several nicknames. For example, in the \textit{Harry Potter} series, the main antagonist, \textex{Tom Marvolo Riddle}, is known under various nicknames: \textex{Lord Voldemort}, \textex{He-Who-Must-Not-Be-Named}, the \textex{Dark Lord}, and others. We annotate all significant nicknames:
\begin{example}
    Rejoice, for \entCHR{You-Know-Who} has gone at last!
\end{example}

It is worth stressing that some characters are referred to using \textit{only} their nicknames, so discarding these would mean missing these characters entirely. In \textit{The Black Company}, one of the main characters is called the \textex{Captain}, and his true name is \textit{never} revealed: 
\begin{example}
    \entLOCwm{Beryl} had ground our spirits down, but had left none so disillusioned as the \entCHR{Captain}.
\end{example}

%%%
\paragraph{Attributes}
When an attribute follows the name of the character, we treat the whole expression as a nickname, as in this example from J. R. R. Tolkien's \textit{The Lord of the Rings}:
\begin{example}
    It has seldom been heard of that \entCHR{Gandalf the Grey} sought for aid \textelp{}
\end{example}
\noindent The attribute is sometimes itself the name of a distinct entity, so this is consistent with our decision to annotate the outmost entity in case of nested entities (cf. Section~\ref{sec:GralNested}). Here are some examples from \textit{A Song of Ice and Fire}:
\begin{example}
    \begin{itemize}[leftmargin=*,itemsep=1mm]
        \item The one I want is with a highborn girl, the daughter of \entCHR{Lord Stark of Winterfell}.
        \item This is the will and word of \entCHR{Robert of House Baratheon}, the First of his Name \textelp{} 
    \end{itemize}
\end{example}
\noindent Note that in the above examples, we do not annotate \textex{Winterfell} as a location or \textex{House Baratheon} as an organization.

%%%
\paragraph{Frequency}
We do not annotate very punctual nicknames or insults (even personalized ones). In the following example from \textit{A Song of Ice and Fire}, we would annotate only the word \textex{Arya}: 
\begin{example}
    \entCHRwm{Jeyne} used to call her \entNOT{Arya Horseface}, and neigh whenever she came near.
\end{example}
\noindent And in this example from \textit{Harry Potter}, only \textex{Potter}:
\begin{example}
    ``\entNOT{Saint Potter}, the \entGRPwm{Mudbloods}' friend,” said \entCHRwm{Malfoy} slowly.
\end{example}
\noindent In this sentence from \textit{The Black Company}, \textex{Goblin} uses a creative nickname to provoke his friend \textex{One-Eyed}:
\begin{example}
    \entCHRwm{Goblin} chortled, ``You ain't winning even when you deal, \entNOT{Maggot Lips}. \textelp''
\end{example}
\noindent Here, we would not annotate any part of the expression \textex{Maggot Lips}, which is used only once in the whole book.

%%%%%%%%%%%%%%%%%%%%%%%%%
\subsection{Societal Roles}
\label{sec:ChrRoles}

%%%
\paragraph{General Rule}
We annotate societal roles according to the same general principles as before, i.e. when they refer to a specific character without ambiguity, and they are mentioned frequently enough. Put differently, we consider them a bit as if they were nicknames.

%%%
\paragraph{Examples}
For instance, in Robin Hobb's \textit{Farseer Trilogy}, \textex{King Shrewd} is the grandfather of the protagonist, and an important character. Moreover, he is the only king for most of the first book, therefore we annotate as follows:
\begin{example}
    But our father the \entCHR{King} is not a hasty man, as well we know.
\end{example}
On the contrary, some roles are too generic or too common to be annotated, as they do not ensure the unicity of the entity:
\begin{example}
    He pointed, and \entCHRwm{Arya} saw it. The body of the \entNOT{soldier}, shapeless and swollen.
\end{example}

%%%
\paragraph{Capitalization}
Capitals are a good indication to detect important societal roles, however many words are capitalized without having such meaning. Moreover, the use of capitals varies significantly from one author to the other. For instance, in \textit{Brave New World}, Aldous Huxley capitalizes a lot of expressions:  
\begin{example}
    The \entNOT{Chief Bottler}, the \entNOT{Director of Predestination}, 3 \entNOT{Deputy Assistant Fertilizer-Generals} %\textelp{}
\end{example}
\noindent Each one of these three expressions appears only once or twice in the whole novel, so we do not consider them as entities.

%%%%%%%%%%%%%%%%%%%%%%%%%
\subsection{Personification}
\label{sec:ChrPerso}
%%%
\paragraph{General Rule}
As per our wide definition of what a character is, we annotate personified animals or items as \texttt{CHR} when relevant. 

%%%
\paragraph{Artificial Beings}
This includes robots and other manufactured beings such as \textex{Marvin the Paranoid Android} from \textit{The Hitchhiker's Guide to the Galaxy} series:
\begin{example}
    \textelp{} \entCHR{Marvin} managed to convey his utter contempt and horror of all things human.
\end{example}
\noindent On the same note, the sentient sword \textex{Stormbringer} in Michael Moorcock's \textit{Cycle of Elric} has its own will, so we annotate it as a \texttt{CHR} entity:
\begin{example}
    \entCHR{Stormbringer} whined almost petulantly, like a dog stopped from biting an intruder.
\end{example}

Cheeses are usually inanimate objects, which makes \textex{Horace the Cheese} a more extreme example of personification. This character appears in the \textit{Discworld} series by Terry Pratchett:
\begin{example}
    \entCHR{Horace} was the only cheese that would eat mice and, if you didn't nail him down, other cheeses.
\end{example}

%%%
\paragraph{Animals}
Lewis Carroll's \textit{Alice in Wonderland} involves many personified animals, such as \textex{Mouse}:
\begin{example}
    \entCHR{Mouse}, do you know the way out of this pool?
\end{example}
\noindent However, we do not annotate common entities without any significant role in the story:
\begin{example}
    \textelp{} and she soon made out that it was only a \entNOT{mouse} \textelp{}
\end{example}

%%%
\paragraph{Abstract Concepts}
Very often, abstract concepts such as fate or death are personified in novels. We annotate them only if they are actual characters. Consider for instance this example from Terry Pratchett's \textit{The Color of Magic}:
\begin{example}
    \entCHR{Death}, insofar as it was possible in a face with no movable features, looked surprised.
\end{example}
\noindent \textex{Death} is a well-known character of the Discworld series, so we annotate him. On the contrary, the following excerpt from \textit{Moby Dick} is a counterexample:
\begin{example}
    Of such a letter, \entNOT{Death} himself might well have been the post-boy.
\end{example}
\noindent This strong personification might suggest that \textex{Death} is a proper character of the novel, but this is not the case, so we do not annotate it.

%%%%%%%%%%%%%%%%%%%%%%%
\subsection{Disjointed Entities}
\label{sec:ChrDisjoint}

%%%
\paragraph{General Rule}
Disjointed names are annotated as characters if each individual entity is self-sufficient, i.e. if the expression referring to this entity is enough to recognize them.

%%%
\paragraph{Examples}
Here is an example from Jane Austen's \textit{Pride \& Prejudice}:
\begin{example}
    \entCHR{Elizabeth}, \entCHR{Kitty} and \entCHR{Lydia Bennet} are sisters.
\end{example}
\noindent In the above sentence, we assume that both \textex{Elizabeth Bennet} and \textex{Kitty Bennet} can be identified by their first names. Even though the family name \textex{Bennet} is implicitly shared by all three mentions, our annotation only associates it to the last character.

This stays true if the shared part of the name is plural, as in this example from Alexandre Dumas' \textit{The Three Musketeers}:
\begin{example}
    \textelp to prevent \entCHR{MM. Bassompierre} and \entCHR{Schomberg} from deserting the army, a separate command had to be given to each.
\end{example}
\noindent Here, \textex{MM.} stands for \textex{Misters} (plural), but we associate it only with \textex{Schomberg} in our annotation.

%%%
\paragraph{Counterexample}
Otherwise, the entire span is annotated as a group entity (cf. Section~\ref{sec:Grp}). See this sentence, also from \textit{Pride \& Prejudice}:
\begin{example}
    \textelp{} \entCHRwm{Mr. Collins}'s scruples of leaving \entGRP{Mr and Mrs Bennet} for a single evening during his visit %\textelp{}
\end{example}
\noindent In the below example, \textex{Mr} is not self-sufficient, so we annotate the whole expression as \texttt{GRP}.

%%%%%%%%%%%%%%%%%%%%%%%%%%%%%%%%%%%%%%%%%%%%%%%%%%%%
\newpage
\section{Location Entities (\texttt{LOC})}
\label{sec:Loc}
We consider that the term \textit{location} denotes physical or metaphysical entities that embody a specific place or region. Locations are devoid of any agency: if an entity is described as performing an active action, it cannot be a \texttt{LOC} entity. In particular, names typical of a location but that refer to geopolitical entities in this context should be annotated as \texttt{ORG} (cf. Section~\ref{sec:ConfusLocOrg}).

%%%%%%%%%%%%%%%%%%%%%%%%%
\subsection{Proper Nouns}
\label{sec:LocProper}

%%%
\paragraph{General Rule}
A number of locations are referred to using a proper noun, in which case we annotate them. 

%%%
\paragraph{Physical Locations}
Physical entities include neighborhoods such as \textex{Flea Bottom} in \textit{A Song of Ice and Fire}, cities such as \textex{London}, regions such as \textex{Derbyshire}, countries such as \textex{England}, continents such as \textex{Westeros} (also from \textit{A Song of Ice and Fire}):
\begin{example}
    \begin{itemize}[leftmargin=*,itemsep=1mm]
        \item She had been sleeping in \entLOC{Flea Bottom}, on rooftops and in stables \textelp{}
        \item \textelp{} I did not feel quite certain that the air of \entLOC{London} would agree with \entCHRwm{Lady Lucas}.
        \item \textelp{} not all his large estate in \entLOC{Derbyshire} could then save him \textelp{}
        \item \textelp{} with a gesture whose significance nobody in \entLOC{England} but the \entCHRwm{Savage} now understood %\textelp{}
        \item Remember, child, this is not the iron dance of \entLOC{Westeros} we are learning \textelp{}
    \end{itemize}
\end{example}
\noindent These examples come from \textit{A Song of Ice and Fire} (\#1, \#5), \textit{Pride and Prejudice} (\#2, \#3), and \textit{Brave New World} (\#4).

Physical locations also include man-made structures such as buildings like \textex{Harrenhal}, a fortress from \textit{A Song of Ice and Fire}: 
\begin{example}
    \textelp{} now he's marching north toward \entLOC{Harrenhal}, burning as he goes.
\end{example}
\noindent There are also commercial buildings like the \textex{Cattery}, a brothel in \textit{A Song of Ice and Fire}:
\begin{example}
    \textelp{} and the brothel called the \entLOC{Cattery}, where he got strange looks but no help.
\end{example}

Physical locations can refer to natural structures or areas, e.g. \textex{Blackwater Bay} in \textit{A Song of Ice and Fire}:
\begin{example}
    \entLOC{Blackwater Bay} was rough and choppy, whitecaps everywhere.
\end{example}
Similarly to what we do for characters' titles (cf. Section~\ref{sec:ChrTitles}), we include qualifiers in the annotation, such as \textex{Bay} in the previous example, or \textex{Southern} in the following excerpt of \textit{The Black Company}:
 \begin{example}
    \entLOC{Southern Forsberg} remained deceptively peaceful.
\end{example}

Finally, stars and planets can also be considered as locations. See this example from Douglas Adams' \textit{Hitchhiker's Guide To The Galaxy}:
\begin{example}
    \textelp{} \entCHRwm{Ford Prefect} was in fact from a small planet somewhere in the vicinity of \entLOC{Betelgeuse}.
\end{example}
In certain cases, the celestial object is not a place, though, so context must be considered. For instance, in \textit{Moby Dick}:
\begin{example}
    What a fine frosty night; how \entNOT{Orion} glitters; what northern lights!
    % thrown among people as strange to him as though he were in the planet \entLOC{Jupiter} % example from Moby Dick
\end{example}
\noindent Here, \textex{Orion} is just a light in the sky. Similarly, we would not annotate the sun or the moon as locations, unless they are used as such.

%%%
\paragraph{Metaphysical Locations}
Metaphysical entities can be very diverse in nature. Some good examples are the \textex{L-Space} from the \textit{Discworld} series, which is a place connecting all libraries across time and space:
\begin{example}
    All libraries everywhere are connected in \entLOC{L-space}. All libraries. Everywhere.
\end{example}
\noindent The warrens from the \textit{Malazan Book of the Fallen}, such as \textex{Omtose Phellack}, are some sort of pocket worlds, and could also be considered as metaphysical locations. Those are simultaneously places that connect with the physical plane, and the source of magic in this lore: 
\begin{example}
    My Warren touches \entLOC{Omtose Phellack}. I can reach it, \entCHRwm{Adjunct}. Any \entGRPwm{T'lan Imass} could.
\end{example}

% what about hell and heaven in interjections (ex. "For heaven's sake!")? they do not mean a place. But we annotate God in "For God's sake" as a character.

%%%%%%%%%%%%%%%%%%%%%%%%%
\subsection{Parts of Names}
\label{sec:LocPart}

%%%
\paragraph{General Rule}
In accordance to our general principle from Section~\ref{sec:GralParts}, we annotate parts of location names, under certain conditions. This is analogous to what we do with characters' titles (cf. Section~\ref{sec:ChrTitles}) and societal roles (Section~\ref{sec:ChrRoles}).

%%%
% \paragraph{Example}
% Je n'ai pas trouvé d'autre type de situation (que les Noun Modifiers). 
% >> Si on tombe sur un autre cas : à compléter ici.

%%%
\paragraph{Noun Modifiers}
Certain location names are constituted of a common noun, acting as a noun modifier, and a proper noun. It is common for these locations to be referred to using only the former. For instance, the previously mentioned \textex{Vale of Arryn}, from \textit{A Song of Ice and Fire}, is frequently referred to simply as the \textex{Vale}:
\begin{example}
    ``A pity \entCHRwm{Lysa} carried them off to the \entLOC{Vale},'' \entCHRwm{Ned} said dryly.
\end{example}
\noindent Similarly, in \textit{Moby Dick}, the Massachusetts island named \textex{Martha's Vineyard} is often simply called the \textex{Vineyard}:
\begin{example}
    \textelp{} once the bravest boat-header out of all \entLOCwm{Nantucket} and the \entLOC{Vineyard}; \textelp{}
\end{example}

%%%
\paragraph{Exceptions}
It is important that the short form allows to uniquely identify the entity, and that it is frequently used. In the following example from \textit{The Black Company}, \textex{avenue} is used only thrice in the book, and to refer to two distinct avenues, so it should not be annotated when used by itself:
\begin{example}
   ``We had come to the \entLOCwm{Avenue of the Syndics'}, \textelp{} There was a procession on the \entNOT{Avenue}.''
\end{example}

Similarly, in this sentence from \textit{A Song of Ice and Fire}, \textex{bay} refers to \textex{Blackwater Bay}:
\begin{example}
    \textelp{} who would stand out in the \entNOT{bay} in case the \entGRPwm{Lannisters} had other ships hidden \textelp{}
\end{example}
\noindent There are several other bays mentioned in the novels, so we do not annotate this word when used separately.

%%%%%%%%%%%%%%%%%%%%%%%%%
\subsection{Definite Descriptions}
\label{sec:LocDescr}

%%%
\paragraph{General Rule}
As for other type of entities, we do not annotate definite descriptions, unless they have an important role in the story.

%%%
\paragraph{Examples}
Consider the following sentence:
\begin{example}
    I am going to \entNOT{the lake}, I'll be back late in the evening.
\end{example}
\noindent If this lake was the only lake mentioned in the novel, and if it was central to the story, then we would annotate it as \texttt{LOC}. A good example is \textex{the Wall} in \textit{A Song of Ice and Fire}, a monumental ice and rock structure spanning hundreds of kilometers:
\begin{example}
    There's not been a direwolf sighted south of the \entLOC{Wall} in two hundred years.
\end{example}

%%%
\paragraph{Numerical Expressions}
Certain expressions include numerical values: we annotate them too, due to the unicity they entail. For instance, from \textit{Brave New World}:
\begin{example}
    \begin{itemize}[leftmargin=*,itemsep=1mm]
        \item Told them of the test for sex carried out in the neighborhood of \entLOC{Metre 200}.
        \item Their wanderings \textelp{} had brought them to the neighborhood of \entLOC{Metre 170} on \entLOC{Rack 9}.
    \end{itemize}
\end{example}

%%%%%%%%%%%%%%%%%%%%%%%
\subsection{Nicknames}
\label{sec:LocNick}

%%%
\paragraph{General Rule}
Although this is not as common as for characters, some locations are also referred to using nicknames. Like before, we annotate them if they are frequently used, and allow identifying the entity reasonably well.

%%%
\paragraph{Example}
Using the \textex{Big Apple} instead of \textex{New York} is a good example:
\begin{example}
    After years of dreaming, she finally arrived in the \entLOC{Big Apple}, ready to pursue her acting career.
\end{example}

%%%%%%%%%%%%%%%%%%%%%%%
% \subsection{Disjoint Names}
% \label{sec:LocDisjoint}
% a priori, pas d'exemple pertinent qui illustrait une situation analogue à ce qu'on peut rencontrer pour les noms de personnages

%%%%%%%%%%%%%%%%%%%%%%%%%%%%%%%%%%%%%%%%%%%%%%%%%%%%
\newpage
\section{Organization Entities (\texttt{ORG})}
\label{sec:Org}
We consider that an organization is an \textit{institutional} entity: a state, a ministry, a guild\ldots{}. By comparison, informal groups such as families, demonyms, or ethnonyms, are annotated as groups instead (see Section~\ref{sec:Grp}).

%%%%%%%%%%%%%%%%%%%%%%%
\subsection{Proper Nouns}
\label{sec:OrgProper}

%%%
\paragraph{General Rule}
As for the other entity types, we annotate all proper nouns referring to organizations.

%%%
\paragraph{Examples}
For organizations, proper nouns are not as common in novels as for characters and locations. For instance:
\begin{example}
    \textelp{} \entORG{Canonical} announced the release of their latest \entMSCwm{Ubuntu} update, promising new features.
\end{example}
\noindent Or the \textex{RAMJAC corporation}, taken from Kurt Vonnegut's \textit{Jailbird}: 
\begin{example}
    That agency \textelp{} is now a wholly-owned subsidiary of The \entORG{RAMJAC Corporation}.
\end{example}
\noindent Here, we include \textex{Corporation} in the annotation, similarly to what we do with honorific titles and qualifiers for other entity types (see Sections~\ref{sec:ChrTitles} and~\ref{sec:LocPart}, for instance).

The Hogwarts houses from \textit{Harry Potter} are also a good example of organizations possessing a proper noun:
\begin{example}
    He took off the hat and walked shakily towards the \entORG{Gryffindor} table.
\end{example}
\noindent Although in this case, there is also a metonymy, as \textex{Godric Gryffindor} is the founder of this house.

%%%
\paragraph{Groups}
The difference with \texttt{GRP} entities is not always obvious: the annotator must take into account the informal vs. institutional nature of the entity, as explained in Section~\ref{sec:ConfusOrgGrp}.

%%%%%%%%%%%%%%%%%%%%%%%
\subsection{Definite Descriptions}
\label{sec:OrgDescr}

%%%
\paragraph{General Rule}
Very often, organizations in novels are referred to only with definite descriptions. Of course, we annotate these expressions, otherwise we would miss completely the corresponding entities.

%%%
\paragraph{Examples}
Here are two examples taken from \textit{A Song of Ice and Fire}:
\begin{example}
    \begin{itemize}[leftmargin=*,itemsep=1mm]
        \item \textelp{} I could sweep the \entORG{Seven Kingdoms} with ten thousand \entGRPwm{Dothraki} screamers.
        % \item \entLOCwm{Storm's End} belonged to \entORG{House Baratheon} for three hundred years \textelp{} % >> exemple pas terrible car Baratheon est un nom propre, donc on est plus sur du noun modifier
        \item \entCHRwm{Theon} is the rightful heir, unless he's dead\ldots{} but \entCHRwm{Victarion} commands the \entORG{Iron Fleet}. 
    \end{itemize}
\end{example}
\noindent And another example from George Orwell's \textit{1984}:
\begin{example}
    The \entORG{Ministry of Peace} concerns itself with war \textelp{}, the \entORG{Ministry of Love} with torture \textelp{}
\end{example}

%%%%%%%%%%%%%%%%%%%%%%%
\subsection{Disjointed Entities}
\label{sec:OrgDisjoint}

%%%
\paragraph{General Rule}
Like for characters (Section~\ref{sec:ChrDisjoint}), disjointed names are annotated as organizations if each individual entity mention is self-sufficient. 

%%%
\paragraph{Example}
Here is an example from \textit{The Blade Itself}:
\begin{example}
    \textelp{} from \textelp{} high-born nobodies to the great magnates of the \entNOT{Open} and \entORG{Closed Councils}.
\end{example}
\noindent The sentence mentions two institutions: the \textex{Open Council} and the \textex{Closed Council}. Only the latter can be recognized in the text: \textex{Open} is not enough to identify the former.

%%%
\paragraph{Counterexamples}
Unlike for characters, it is rarely the case that each individual entity mention is self-sufficient, because the common portion of the entity name is often necessary to recognize the organization. 

See for instance this example from \textit{Brave New World}:
\begin{example}
    Then came the \entORG{Bureaux of Propaganda by Television}, \entNOT{by Feeling Picture}, and \entNOT{by Synthetic Voice and Music} respectively–twenty-two floors of them. %\textelp{}
\end{example}
\noindent This sentence lists three distinct organizations: the \textex{Bureau of Propaganda by Television}, the \textex{Bureau of Propaganda by Feeling Picture}, and the \textex{Bureau of Propaganda by Synthetic Voice and Music}. However, only the first mention is recognizable by itself, hence our single annotation.

Another example, this time from \textit{The Blade Itself}:
\begin{example}
    He established the \entNOT{Councils}, \entNOT{Closed} and \entNOT{Open}, he formed the \entORGwm{Inquisition}.
\end{example}
\noindent This sentence refers to character \textex{Bayaz} creating three institutions: the \textex{Closed Council}, the \textex{Open Council}, and the \textex{Inquisition}. The councils are not recognizable by using only the \textex{Closed} and \textex{Open} parts of their full names.

%%%%%%%%%%%%%%%%%%%%%%%%%%%%%%%%%%%%%%%%%%%%%%%%%%%%
\newpage
\section{Group Entities (\texttt{GRP})}
\label{sec:Grp}
We define group entities as informal gathering or sets of characters, that do not have any proper institutional existence. They are used when the concerned name does not refer to an individual character but several ones at once, while still providing sufficient information to be able to identify them relatively well.

The rationale for annotating groups is that some authors extract character networks that contain vertices representing such groups. For instance, when studying Homer's \textit{Iliad}, Venturini \textit{et al}.~\cite{Venturini2016} model certain Greek tribes using a single vertex (e.g. \textex{Myrmidons}), while in~\cite{Falk2016}, Falk represents bystanders collectively, using a specific vertex.

%%%%%%%%%%%%%%%%%%%%%%%%%
\subsection{Family Names}
\label{sec:GrpFamily}

%%%
\paragraph{General Rule}
We annotate family names as \texttt{GRP} entities when they refer to several members of that family. 

%%%
\paragraph{Example}
In the below example, \textex{Baggins} is a family name from \textit{The Lord of the Rings}, and it is used to refer to the family as a whole:
\begin{example}
    But there you are: \entGRPwm{Hobbits} must stick together, and especially \entGRP{Bagginses}.
\end{example}

%%%%%%%%%%%%%%%%%%%%%%%%%
\subsection{Demonyms \& Ethnonyms}
\label{sec:GrpDemonyms}

%%%
\paragraph{General Rule}
We annotate ethnonyms (names referring to ethnic groups) and demonyms (names referring to the inhabitants of a place) as \texttt{GRP}. 

In the following example, \textex{Chyurda} is the name of people living in a specific kingdom, in the \textit{The Farseer Trilogy}:
\begin{example}
    That was the first year the \entGRP{Chyurda} tried to close the pass.
\end{example}
\noindent The novel \textit{Brave New World} by Aldous Huxley provides another good example, with the caste system it describes:
\begin{example}
    We decant our babies as socialized human beings, as \entGRP{Alphas} or \entGRP{Epsilons}.
\end{example}

%%%
\paragraph{Adjectives}
We also annotate demonyms and ethnonyms when used as adjectives. For instance, \textex{Dothraki} are an ethnic group of nomadic warrior in \textit{A Song of Ice and Fire}:
\begin{example}
    \begin{itemize}[leftmargin=*,itemsep=1mm]
        \item ``If the \entCHRwm{beggar king} crosses with a \entGRP{Dothraki} horde at his back, the traitors will join him.''
        \item A \entGRP{Dothraki} wedding without at least three deaths is deemed a dull affair.
    \end{itemize}
\end{example}
\noindent In the first sentence, the word \textex{Dothraki} explicitly refers to people. It is not the case in the second one, where it rather refers to the Dothraki \textit{culture}. To keep our annotation rules simple, we annotate it nevertheless.

Here is a (extreme) limit case from \textit{Brave New World}:
\begin{example}
    In a little grassy bay between tall clumps of \entGRP{Mediterranean} heather \textelp{}
\end{example}
\noindent As before, for the sake of simplicity, we consider that \textex{Mediterranean} is an adjective derived from a demonym.

In the following example from \textit{The Black Company}, \textex{Arctic} is a bit tricky:
\begin{example}
    \entNOT{Arctic} imps giggled and blew their frigid breath through chinks in the walls of my quarters.
\end{example}
\noindent Here, \textex{Arctic} means \textit{northern}: it is not a denomym, as there is no Arctic continent or people in this fantasy world.

%%%
\paragraph{Languages}
The same word is often used to refer to a social group and to the language of its people. However, note that we annotate languages as cultural objects, see Section~\ref{sec:MscWorks}.

%%%%%%%%%%%%%%%%%%%%%%%%%
\subsection{Definite Descriptions}
\label{sec:GrpDescr}

%%%
\paragraph{General Rule}
A number of groups of characters are described using definite descriptions. We annotate them as \textex{GRP} provided they exhibit the usual properties of capitalization, frequency and unicity. This type of groups is sometimes difficult to distinguish from organizations: see Section~\ref{sec:ConfusOrgGrp} for more detail on this topic.

%%%
\paragraph{Example}
In the following example from Suzanne Collins's \textit{The Hunger Games} series, the expression \textex{Career Tributes} denotes a set of characters that are grouped because of one of their attribute, without any institutional existence or structure. Therefore, we annotate it as \texttt{GRP}:
\begin{example}
    In \entLOCwm{district 12}, we call them the \entGRP{Career Tributes} \textelp{}
\end{example}

%%%
\paragraph{Counterexample}
In \textit{Brave New World}, Aldous Huxley likes to refer to groups of people through their job or role in the society, and capitalizes the expression: 
\begin{example}
    Bent over their instruments, three hundred \entNOT{Fertilizers} were plunged \textelp{}
\end{example}
\noindent We do not annotate such expressions, unless they are frequently mentioned as a group. Here, \textex{Fertilizers} appear only once in the whole novel.

% faut trouver d'autres exemples pour ici

%%%%%%%%%%%%%%%%%%%%%%%%%
\subsection{Enumerations}
\label{sec:GrpEnums}

%%%
\paragraph{General Rule}
We annotate enumerations of multiple entities as groups if these entities are not self-sufficient. 

%%%
\paragraph{Example}
In the following example, \textex{Mr} is not self-sufficient, so we annotate the whole expression as \texttt{GRP}:
\begin{example}
    \entGRP{Mr and Mrs Bennet} plan to go to \entLOCwm{London} soon.
\end{example}

%%%
\paragraph{Example}
On the contrary, in the following example from \textit{The Three Musketeers}, each individual character is clearly identified, and therefore annotated separtely: 
\begin{example}
    \textelp{} highly applauded, except by \entCHR{MM. Grimaud}, \entCHR{Bazin}, \entCHR{Mousqueton}, and \entCHR{Planchet}.
\end{example}

%%%%%%%%%%%%%%%%%%%%%%%%%%%%%%%%%%%%%%%%%%%%%%%%%%%%
\newpage
\section{Miscellaneous Entities (\texttt{MSC})}
\label{sec:Msc}
This category gathers various types of entities likely to be of interest.

%%%%%%%%%%%%%%%%%%%%%%%%%
\subsection{Temporal Entities}
\label{sec:MscEvts}

%%%
\paragraph{General Rule}
We annotate named temporal entities, also called praxonyms, as \texttt{MSC}. This encompasses events such as revolutions, crises, festivals, etc., and as well as historical periods.

%%%
\paragraph{Holidays \& Festivals}
We annotate holidays and festivals, provided they have a name. For instance, in \textit{Moby Dick}:
\begin{example}
    Now, it being \entMSC{Christmas} when the ship shot from out her harbor \textelp{}
\end{example}

%%%
\paragraph{Events}
Many historical events are rather punctual, compared to historical periods that are discussed below. See this excerpt of \textit{A Song of Ice and Fire}:
\begin{example}
    The \entMSC{Red Wedding} was my father's work, and \entCHRwm{Ryman}'s and \entCHRwm{Lord Bolton}'s.
\end{example}
\noindent The expression \textex{Red Wedding} refers to an event that lasted a few hours, and constitutes a turning point in the story. 

Alternatively, the event can be hypothetical, as in this example from \textit{The Three Musketeers}:
\begin{example}
    at the day of the \entMSC{Last Judgment} \entCHRwm{God} will separate blind executioners from iniquitous judges?
\end{example}

%%%
\paragraph{Periods}
This type of annotation also concerns historical periods. For instance, from the \textit{A Song of Ice and Fire}:
\begin{example}
    \begin{itemize}[leftmargin=*,itemsep=1mm]
        \item ``There was one knight,'' said \entCHRwm{Meera}, ``in the \entMSC{Year of the False Spring} \textelp{}
        \item The signing of the \entMSCwm{Pact} ended the \entMSC{Dawn Age}, and began the \entMSC{Age of Heroes}.
    \end{itemize}
\end{example}
\noindent Both \textex{Dawn Age} and \textex{Age of Heroes} refer to periods in the ancient history of this lore. Similarly, in \textit{The Black Company}:
\begin{example}
    The \entORGwm{Company} was in service to the \entCHRwm{Archon of Bone}, during the \entMSC{Revolt of the Chiliarchs}.
\end{example}

Sometimes, the distinction between punctual event and period is not obvious. For instance, in \textit{A Song of Ice and Fire}:
\begin{example}
    ``They date from before \entMSC{Aegon's Conquest},'' \entCHRwm{Cersei} explained to her.
\end{example}
\noindent The conquest of \textex{Westeros} by \textex{Aegon} took some time, so technically it is a period. But here, the expression \textex{Aegon's Conquest} actually refers to his Aegon's \textit{coronation}, which is used as a reference date by the historians in this lore (akin to BC/AD in the real world).

%%%
\paragraph{Dates}
We do not annotate dates in general:
\begin{example}
    \begin{itemize}[leftmargin=*,itemsep=1mm]
        \item D'you know what that little girl of mine did last \entNOT{Saturday}, when her troop was on a hike \textelp{} %? % 1984
        \item Last \entNOT{Monday} (\entNOT{July 31st}) we were nearly surrounded by ice \textelp{} % Frankenstein
        \item \textelp{} the question didn't arise; in this year of stability, \entNOT{A.F. 632}, it didn't occur to you %to ask it. % Brave New World
    \end{itemize}
\end{example}
\noindent The first and second examples are from \textit{1984} and Mary Shelley's \textit{Frankestein}, whose worlds and calendars are similar to ours. The third one is from \textit{Brave New World}, in which dates are expressed relative to the production of the first \textit{Ford T} automobile, hence the \textex{A.F.} (\textit{Anno Ford}).

% example of explicit date in Moby Dick:
% [...] was lost overboard, Near the Isle of Desolation, off Patagonia, November 1st, 1836. 
% (not particularly important for the story, though)

One justification for not annotating dates is that they are usually considered as a separate specific entity for NER. Moreover, there are tools specifically designed to handle them, such as HeidelTime~\cite{Stroetgen2015}.

% TODO
% One exception is if the date is significant importance in the novel, e.g.:
% j'ai essayé de demander à ChatGPT mais il me renvoie des évènements nommées :< - AA
% ...

%%%%%%%%%%%%%%%%%%%%%%%%%
\subsection{Cultural Assets}
\label{sec:MscWorks}

%%%
\paragraph{General Rule}
We annotate as \texttt{MSC} important artistic and intellectual works, as well as cultural objects.

%%%
\paragraph{Intellectual and Artistic Works}
Intellectual works include books, songs, movies, paintings, etc. We annotate their names or titles when they appear explicitly in the text, for instance:
\begin{example}
    \begin{itemize}[leftmargin=*,itemsep=1mm]
        \item We will take this book, the \entMSC{Book of Mazarbul}, and look at it more closely later.
        \item \textelp{} like the men singing the \entMSC{Corn Song}, beautiful, beautiful, so that you cried \textelp{}
        \item \textelp{} to which \entCHRwm{Helrnholtz} had recently been elected under \entMSC{Rule Two}.
    \end{itemize}
\end{example}
\noindent The first example come from \textit{The Lord of the Rings}, and both others from \textit{Brave New World}.

%%%
\paragraph{Cultural Objects}
Cultural objects encompass dishes and wines, as in this example from \textit{A Song of Ice and Fire}:
\begin{example}
    There is a flagon of good \entMSC{Arbor gold} on the sideboard, \entCHRwm{Sansa}.
\end{example}
\noindent We can also mention spirits, for instance, in \textit{Moby Dick}:
\begin{example}
     \textelp{} and with a benevolent, consolatory glance hands him--what? Some hot \entMSC{Cognac}?
\end{example}

There are also games, such as this fictional card game from \textit{The Black Company}:
\begin{example}
    We were playing head-to-head \entMSC{Tonk}, a dull time-killer of a game.
\end{example}
\noindent And sports, like in this excerpt of \textit{Brave New World}:
\begin{example}
    The crowds that daily left \entLOCwm{London}, left it only to play \entMSC{Electromagnetic Golf} or \entMSC{Tennis}.
\end{example}

Cultural objects also include a wide array of similar concepts, e.g. a scientific technique in \textit{Brave New World}:
\begin{example}
    But \entMSC{Podsnap's Technique} had immensely accelerated the process of ripening.
\end{example}
\noindent Or a commercial brand in \textit{Harry Potter}:
\begin{example}
    He had patched up his wand with some borrowed \entMSC{Spellotape} \textelp{} %, but it seemed to be damaged beyond repair.
\end{example}
\noindent Or the motto of the noble houses in \textit{A Song of Ice and Fire}:
\begin{example}
    All but the \entGRPwm{Starks}. `Winter is coming,' said the \entMSC{Stark words}.
\end{example}
% \noindent Or a motto, such as the \textex{World State} motto in \textit{Brave New World}:
% \begin{example}
%     He quoted the planetary motto. ``\entMSC{Community, Identity, Stability}.'' Grand words.
% \end{example}

%%%
\paragraph{Languages}
We include languages in this category. This is a bit far-fetched, but it allows distinguishing languages from demonyms, which often take the exact same form in English.

Here is an example from \textit{The Three Musketeers}:
\begin{example}
    \entCHRwm{D'Artagnan} did not know \entLOCwm{London}; he did not know a word of \entMSC{English} \textelp{} %; but he wrote the name of \entCHRwm{Buckingham} on a piece of paper
\end{example}

We annotate fictional languages too, such as in this example from \textit{A Song of Ice and Fire}:
\begin{example}
    They had no common language. \entMSC{Dothraki} was incomprehensible to her \textelp{}
\end{example}

%%%%%%%%%%%%%%%%%%%%%%%%%
\subsection{Named Artefacts}
\label{sec:MscItems}

%%%
\paragraph{General Rule}
We annotate named items, also called ergonyms, as \texttt{MSC}, provided they are \textit{not} sentient.

%%%
\paragraph{Example}
For instance, \textex{King Arthur}'s sword \textex{Excalibur} is magical, but does not act independently nor communicate:
\begin{example}
    \entCHRwm{Arthur} drew his sword \entMSC{Excalibur} that he had gained by \entCHRwm{Merlin} from \entCHRwm{Vivian}.
\end{example}

Many vehicles are also named, and are consequently annotated as \texttt{MSC}, e.g.:
\begin{example}
    \begin{itemize}[leftmargin=*,itemsep=1mm]
        \item \entCHRwm{Batman} raced through \entLOCwm{Gotham City} streets in the \entMSC{Batmobile}, ready for action.
        \item \entMSC{Black Betha} rode the flood tide, her sail cracking and snapping at each shift of wind.
    \end{itemize}
\end{example}
\noindent The first example is invented, the second comes from \textit{A Song of Ice and Fire}.

%%%
\paragraph{Counterexample}
However, as explained in Section~\ref{sec:ChrPerso}, named items that are sentient are annotated as characters. This is the case, for instance, of \textex{Elric}'s sword \textex{Stormbringer}, or of the \textex{Sorting Hat} in \textit{Harry Potter}:
\begin{example}
    The \entMSC{Sorting Hat} chose you for \entORGwm{Gryffindor}, didn't it? And where's \entCHRwm{Malfoy}?
\end{example}

%%%%%%%%%%%%%%%%%%%%%%%%%
\subsection{Other Entities}
\label{sec:MscOthers}
A number of other kinds of entities are referred to by proper names in novels, in which case we annotate them as miscellaneous entities.

%%%
\paragraph{Meteorological Events}
These events, also called \textit{phenonyms}, include tempests, cyclones, etc. Here is an example from \textit{Moby Dick} that mentions the name of a wind:
\begin{example}
    \textelp{} where that tempestuous wind \entMSC{Euroclydon} kept up a worse howling than ever it did \textelp{}
\end{example}

%%%
\paragraph{Awards \& Decorations}
We annotate the names of awards and decoration, as in this example from George Orwell's \textit{1984}: 
\begin{example}
    \entCHRwm{Comrade Withers} \textelp{} had been \textelp{} awarded a decoration, the \entMSC{Order of Conspicuous Merit} %\textelp{}
\end{example}

%%%%%%%%%%%%%%%%%%%%%%%%%%%%%%%%%%%%%%%%%%%%%%%%%%%%
\newpage
\section{Type Confusion}
\label{sec:Confus}
In certain cases, it is not clear what the type of an entity is. In this section, we focus on this issue, and provide some example that aim at helping to make such distinction.

%%%%%%%%%%%%%%%%%%%%%%%%%
\subsection{Characters vs. Locations}
\label{sec:ConfusChrLoc}

%%%
\paragraph{Characters as Locations}
Some organizations locations are named after a person's name. For instance, a commercial building such as \textex{Morrogo's}, an inn in \textit{A Song of Ice and Fire}, is named after its owner:
\begin{example}
    \entCHRwm{Sam} began his search at \textelp{} \entLOC{Moroggo's}, places where \entCHRwm{Dareon} had played before.
\end{example}
\noindent Here, including the genitive \textex{'s} in the annotation is consistent with our decision to annotate the outmost entity in nested entities (cf. Section~\ref{sec:GralNested}). Here is another example, also from \textit{A Song of Ice and Fire}:
\begin{example}
    \textelp{} some stranger from the \entLOC{Vale of Arryn} whose name she had forgotten \textelp{}
\end{example}
\noindent In this case, \textex{Arryn} is the name of the person that claimed this territory after a battle. When the name contains a qualifier, such as \textex{Vale} here, the distinction between character and location is much clearer.

The name of the person may be used without any modification, e.g.
\begin{example}
    \entLOCwm{London Zoo} is approximately a 30 minute walk from \entLOC{Saint Pancras}.
\end{example}
\noindent Where \textex{Saint Pancras} is a train station, and not the person \textex{Pancras of Rome}.

%%%
\paragraph{Locations as Characters}
It happens that a location strongly associated with a character is used in place of their name. We annotate the location name as a \texttt{CHR}. See, for instance, this excerpt from \textit{Moby Dick}:
\begin{example}
    It drags hard; I guess he's holding on. Jerk him, \entCHR{Tahiti}! Jerk him off; we haul in no cowards here.
\end{example}
\noindent One of the unnamed seamen is from \textex{Tahiti}, and \textex{Captain Ahab} uses the name of this location instead of the proper character's name. 

%%%
\paragraph{Double Meaning}
Sometimes, the confusion between character and location is more conceptual, as a large entity can be both a character and a place. For instance, in \textex{The Adventures of Pinocchio}, \textex{Pinocchio} and \textex{Geppetto} are swallowed by a giant \textex{shark}, which appears first as a character in the story, before becoming a place where \textex{Pinocchio} can have a walk:
\begin{example}
--- ``Is this \entCHR{Shark} that has swallowed us very long?'' asked the \entCHRwm{Marionette}. \\
--- ``His body, not counting the tail, is almost a mile long.'' \textelp{} \\
\entCHRwm{Pinocchio} \textelp{} began to walk as well as he could inside the \entLOC{Shark}, toward the faint light which glowed in the distance.
\end{example}
\noindent Another similar example is \textex{Erythro}, the sentient planet at the center of Isaac Asimov's novel \textit{Nemesis}:
\begin{example}
    \begin{itemize}[leftmargin=*,itemsep=1mm]
        \item \textelp{} her thoughts were often on \entLOC{Erythro}, the planet they had been orbiting almost all her life.
        \item \entCHR{Erythro} had knowledge of only one kind of mind--its own.
    \end{itemize}
\end{example}

%%%%%%%%%%%%%%%%%%%%%%%%%
\subsection{Characters vs. Organizations}
\label{sec:ConfusChrOrg}

%%%
\paragraph{Characters as Organizations}
Metonymy is quite frequent when referring to organizations, which can lead to a certain confusion. In the below examples from \textit{The Black Company}, it is the case between the person nicknamed \textex{White Rose} and her armed group:
\begin{example}
    \begin{itemize}[leftmargin=*,itemsep=1mm]
     \item I told him about the \entCHR{White Rose}, the lady general who had brought the \entORGwm{Domination} down %\textelp{}
     \item The \entCHRwm{Lady} is no exception. The \entORG{Sons of the White Rose} are everywhere.
     \item \entORG{White Rose} prophets and \entORGwm{Rebel} mainforcers were the least of our troubles.
    \end{itemize}
\end{example}
\noindent The full name of this organization is \textex{Sons of the White Rose}, but the expression \textex{White Rose} is more frequently used as a shortened form.

%%%%%%%%%%%%%%%%%%%%%%%%%
\subsection{Characters vs. Miscellaneous}
\label{sec:ConfusChrMsc}

%%%
\paragraph{Characters as Items}
Metonymy between persons and the various objects covered by the \texttt{MSC} type are quite frequent. Consider these examples from \textit{Moby Dick}: 
\begin{example}
    \begin{itemize}
        \item \entCHRwm{Charity}, his sister, had placed a small choice copy of \entMSC{Watts} in each seaman's berth.
        \item \textelp{} seemed made of solid bronze, \textelp{} like \entCHRwm{Cellini}'s cast \entMSC{Perseus}.
    \end{itemize}
\end{example}
\noindent Isaac Watts is an author of religious hymns, and in the first sentence, \textex{Watts} refers to one of his books. In the second sentence, \textex{Cellini} is the sculptor Benvenuto Cellini, and \textex{Perseus} does not refer directly to the Greek hero, but rather to one of Cellini's work representing the eponymous mythological figure, and titled \textit{Perseus with the Head of Medusa}.

% For instance, in the below examples, the name of an individual is used to refer to the object they designed or fabricated:
% \begin{example}
%     \begin{itemize}[leftmargin=*,itemsep=1mm]
%         \item Violonist \entCHRwm{Niccolò Paganini} is known to have possessed at least one \entMSC{Stradivarius}. % invented
%         \item Don't wait for me, I'll take the \entMSC{Bugatti} to come back. % invented
%     \end{itemize}
% \end{example}
% \noindent In the former example, one could even argue that there are two levels of metonymy: first between \textex{Ettore Bugatti} and its automobile company, second between this company and the cars it produces. 

%%%%%%%%%%%%%%%%%%%%%%%%%
\subsection{Locations vs. Organizations}
\label{sec:ConfusLocOrg}
Metonymy is frequently used between locations and organizations. This can make it difficult to distinguish between \texttt{LOC} and \texttt{ORG} entities, as one name can be used as a location as well as an organization. In both case, it is important to take the context into account in order to decide of the entity type.

%%%
\paragraph{Locations as Organizations}
On the one hand, some location name can be used to refer to an organization. In \textit{A Song of Ice and Fire}, the \textex{Citadel} is the name of a neighborhood hosting the headquarters for the \textex{maesters}, a group of scholars. In the following example, this name is used to refer to the organization instead of the place:
\begin{example}
    Last year when he took ill, the \entORG{Citadel} had sent \entCHRwm{Pylos} out from \entLOCwm{Oldtown} \textelp{} 
\end{example}
\noindent Another example using a country name (\textex{England}) in \textit{Harry Potter}, when the author actually means an organization (the national Quidditch team):
\begin{example}
    \textelp{} he could have played for \entORG{England} if he hadn't gone off chasing dragons.
    % And Neville will play Quidditch for England before Hagrid lets Dumbledore down.
\end{example}

%%%
\paragraph{Organizations as Locations}
On the other hand, sometimes the name of an organization is used to refer to its location. For instance, in \textit{Harry Potter}, \textex{Hogwarts} is the name of an institution, but it is often used to denote the school grounds:
\begin{example}
    \textelp{} he had ten minutes left to get on the train to \entLOC{Hogwarts} and he had no idea how to do it
\end{example}
\noindent Another example, this time from George Orwell's \textit{1984}:
\begin{example}
    A kilometre away the \entLOC{Ministry of Truth}, his place of work, towered vast and white \textelp{}
\end{example}
\noindent The expression \textex{Ministry of Truth} does no refer to the actual organization here, but rather to the building hosting it. Similarly, from the same novel:
\begin{example}
    \entCHRwm{Winston} has never been inside the \entLOC{Ministry of Love}, nor within half a kilometer of it.
\end{example}

%%%
\paragraph{Undetermined}
In cases where the context is unclear about the type of the entity (it could be \texttt{LOC} as well as \texttt{ORG}), we annotate it as \texttt{ORG}. For instance, in \textit{A Song of Ice and Fire}, the \textex{Citadel} sometimes means a place, and sometimes the scholar organization sitting at this place:
\begin{example}
   That is so, my lady. The white ravens fly only from the \entORG{Citadel}.
\end{example}
\noindent In the above example, it is not clear whether \textex{the Citadel} is the place from which the ravens fly, or the organization that send them, or even both of them. In \textit{The Black Company}, we have a similar case:
\begin{example}
   \textelp{} from one of several nearby dives frequented by the \entORG{Bastion} garrison.
\end{example}
\noindent The \textex{Bastion} could as well be the organization localized in this building and commanding the garrison, as the building hosting this garrison.

Also in \textit{The Black Company}, the \textex{Jewel Cities} are a group of geographically and culturally close cities. Since the name does not refer to a group of people, it cannot be annotated as \texttt{GRP}. Depending on the context, the expression can be a location or an organization, but it is not always obvious, e.g.:
\begin{example}
   \entCHRwm{Soulcatcher} commanded the Guard and allies from the \entORG{Jewel Cities}.
\end{example}
\noindent Here, we apply the principle mentioned earlier and opt for \texttt{ORG}.

%%%%%%%%%%%%%%%%%%%%%%%%%
\subsection{Locations vs. Miscelaneous}
\label{sec:ConfusLocMsc}

%%%
\paragraph{Locations as Events}
By metonymy, the name of a location can be used to refer to an important event that took place in this location. In the following example from \textit{The Black Company}, \textex{Beryl} is a city where a battle took place:
\begin{example}
    But \entCHRwm{Soulcatcher} is in high favor since \entMSC{Beryl}, and the \entCHRwm{Limper} isn't because of his failures.
\end{example}

%%%%%%%%%%%%%%%%%%%%%%%%%
\subsection{Organizations vs. Groups}
\label{sec:ConfusOrgGrp}

%%%
\paragraph{Principle}
The distinction between organizations and groups is not always obvious. Size can be a clue, as organizations tend to be larger. Possessing a proper noun would also shift the balance for being an organization. But the main difference is the institutional nature of the entity, i.e. how official it is.

%%%
\paragraph{Examples}
Here are some limit cases. In \textit{Harry Potter}, the \textex{Marauders} are a group of four students that gathered mainly to make mischief:
\begin{example}
    ``Maybe the \entORG{Marauders} never knew the room was there,'' said \entCHRwm{Ron}.
\end{example}
\noindent In this case, the fact that they have their own name would suggest a certain form of recognition. On the one hand, there is no proper structure within the group, but on the other hand, the secret nature of the \textex{Marauders} hints at a certain level of organization. For these reasons, and although this decision is debatable, we annotate it as an \texttt{ORG}.

Also in \textit{Harry Potter}, we annotate \textex{Dudley Dursley's gang} as a group and not an organization, for the exact opposite reasons:
\begin{example}
    \entCHRwm{Harry} was glad school was over, but there was no escaping \entGRP{Dudley's gang} \textelp{}
\end{example}
\noindent It is just a group of children informally gathered by \textex{Harry}'s cousin \textex{Dudley}. Its name is fixed and appears relatively frequently, which is why we consider it as a proper group, and do not annotate only \textex{Dudley} as a character.

In the \textit{Lord of the Rings}, despite the title of the first volume in the trilogy, the \textex{Fellowship of the Ring} is actually called \textex{\textit{Company} of the Ring} in the text:
\begin{example}
    The \entORG{Company of the Ring} stood silent beside the tomb of \entCHRwm{Balin}.
\end{example}
\noindent Whatever its name, this fellowship is formed at a council, and it is constituted of nine members selected to represent specific races from the lore. This is all very organized, which is why we annotate it as an \texttt{ORG}.

%%%%%%%%%%%%%%%%%%%%%%%%%
% \subsection{Groups vs. Miscellaneous}
% \label{sec:ConfusGrpMsc}
% The only case for now is confusing an ethnic group with its language (eg French).

%%%%%%%%%%%%%%%%%%%%%%%%%%%%%%%%%%%%%%%%%%%%%%%%%%%%
\newpage
\section{Concluding Remarks}
\label{sec:Conclu}
% general points
It is worth stressing that having to deal with novels has a significant influence on the annotation process. It is necessary to have a global vision of the whole document to perform this task correctly. This helps to identify elements that are specific to the considered story and to its environment: Who are the characters? What can be considered as frequent in this book? Which honorifics are used in the word of the novel? What are the nicknames? The places? Which conventions does the author use, for instance concerning capitalization?

% work distribution
Concretely, this means that it is more efficient and reliable for one person to annotate the whole book than a few chapters. Also, it is preferable to have a single person annotating the whole book than dividing up the chapters among several persons. Moreover, the annotator should not hesitate to use additional resources. This includes tools such as the \textit{Calibre} ebook viewer\footnote{https://calibre-ebook.com/}, that allow them to assess how frequent some expression is, in order to determine whether it should be annotated as an entity or not. Such tools can also help to determine whether some expression uniquely identifies some entity.

% use wikis
Concerning novels that take place in imaginary worlds, especially Fantasy and Science-Fiction realms, it is particularly important to leverage the available online wikis. These are generally elaborated by fans and are very complete, exploring the lore in detail, providing a Web page for each entity, even minor. A number of concepts in such novels are completely foreign to an unfamiliar reader: a made-up name could as well be a character, a location, an organization, or even a honorific title. Such encyclopedic resources help a lot in alleviating ambiguities, and more generally, making certain annotation decisions.

% take notes
Finally, our last piece of advice is to keep notes of the decisions one makes while annotating a book. For instance, was this group of people annotated as a \texttt{GRP} or an \texttt{ORG}? Was this expression annotated at all? Indeed, a significant amount of text can separate two mentions of the same entity, and one may forget how they previously handled it. Keeping notes helps keeping the annotation process consistent.

%%%%%%%%%%%%%%%%%%%%%%%%%%%%%%%%%%%%%%%%%%%%%%%%%%%%
\appendix
\newpage
\section{Version History}
\label{sec:ApdxVersion}
We use three-part version numbers of the form major--minor--patch for both these guidelines and the \textit{Novelties} corpus. Concretely, \textit{major} changes correspond to very significant modifications of the rules, such as the introduction of a new type of entity. \textit{Minor} changes are modifications of the existing rules (through their edition, addition, deletion). For instance, we could decide to include the determiners in the annotations. Finally, the \textit{patch} level concerns the correction of errors or the clarification of existing rules, for instance by adding new examples to the guidelines, that illustrate cases never met before.

\begin{center}
    \begin{tabularx}{\linewidth}{l l X p{1.3cm}}
        \hline
        \textbf{Version} & \textbf{Date} & \textbf{Changes} & \textbf{Corpus} \\
        \hline
        \texttt{0.1.0} & 25/01/2023 & Beta version based on the work by~\citet{Dekker2019}, with some corrections and additional entity types \texttt{LOC} and \texttt{ORG}. & \texttt{0.1.0}, \texttt{0.2.0} \\ 
        \hline
        \texttt{1.0.0} & 07/03/2024 & Significant revision, based on our feedback from the first annotation pass: more precise definition of a named entity; entity type \texttt{PER} becomes \texttt{CHR}; new entity types \texttt{GRP} and \texttt{MSC} added. & \texttt{1.0.0} \\
        \hline
        \texttt{1.0.1} & 11/06/2024 & Minor revision: languages now considered as \texttt{MSC} entities, instead of being ignored. & \texttt{1.0.0} \\
        \hline
    \end{tabularx}
\end{center}

%%%
\paragraph{Version \texttt{0.1.0}}
This is the beta version of our guidelines. We kick-started our own corpus by leveraging the OWTO corpus (\textit{Out With The Old}) proposed by~\citet{Dekker2019}. Consequently, our very first guidelines are exactly the same as theirs. However, the original annotation guidelines of~\citet{Dekker2019} are extremely minimal, and the OWTO corpus exhibits encoding, tokenization, quoting and annotation problems. We leveraged the experience gained from correcting these errors to modify the guidelines, making them slightly more precise, and adding a few examples. We also extended the scope of entity types, adding \textit{locations} (\texttt{LOC}) and \textit{organizations} (\texttt{ORG}). We followed this version of the guidelines to produce version \texttt{0.1.0} of \textit{Novelties}, which is used in~\cite{Amalvy2023}, and version \texttt{0.2.0}, which was not used in any published work.
%This version of the corpus was introduced and used in~\citet{Amalvy2023}. It consists in a corrected and extended version of the dataset of~\citet{Dekker2019}. The original dataset is composed of one chapter for 40 novels, for which only \texttt{PER} entites are annotated. The selected novels are from classical and modern copyrighted works. Our contributions to enhance this corpus are twofold: first, we fix dataset issues, namely encoding, tokenization, quoting and annotation problems. Second, we enhance the dataset by annotating \texttt{LOC} and \texttt{ORG} entities. The original annotation guidelines of~\citet{Dekker2019} are extremely minimal, so we wrote slightly more precise guidelines with a few examples. In this version, the datasets consists of $4,476$ \texttt{PER}, $886$ \texttt{LOC} and $201$ \texttt{ORG} entities.

%%%
\paragraph{Version \texttt{1.0.0}}
This is a major revision of our guidelines. It is based on our experience in annotating new chapters and even full novels, in an attempt to expand \textit{Novelties}. We released an extensive annotation guide, including a general explanation of our concept of named entity, a list of different annotation cases for each entity type along with positive and negative examples, and a specific part dedicated to possible confusions between types. In terms of major changes in the guidelines themselves, we specifically extended the \texttt{PER} class to include all characters (such as animated weapons, sentient magical creatures, robots\ldots{}), and thus renamed it \texttt{CHR} for \textit{character}. We also introduced two new classes of entities: \textit{groups} (\texttt{GRP}) and \textit{miscellaneous} (\texttt{MSC}). \texttt{MSC} entities are common in other corpora and allow us to annotate additional entities of interest, while \texttt{GRP} entities allow us to distinguish between single characters or groups of them.

%%%
\paragraph{Version \texttt{1.0.1}}
In version \texttt{1.0.0} of the guidelines, we do not annotate languages (French, Dothraki, etc.) at all. This can be a bit confusing for the annotator, considering that the same word is often used in English to refer to a people and to its language. To make things clearer, in this minor revision we change this and annotate languages as cultural assets, using tag \texttt{MSC}. 
%Note that we did not apply this change to the texts previously annotated with version \texttt{1.0.0}, as it is very minor and concern only \textit{miscellaneous} entities.

%%%%%%%%%%%%%%%%%%%%%%%%%%%%%%%%%%%%%%%%%%%%%%%%%%%%
\newpage
\section{Todo List}
\label{sec:ApdxTodo}
Here is a list of examples missing from this document:
\begin{itemize}
    \item Is there an example of explicit date (numbers) that is significant for the novel? (Section~\ref{sec:MscEvts})
    \item Are there some situations where disjoint names are used for locations? (Section~\ref{sec:LocDisjoint})
    \item Is the use of noun modifiers the only situation where only a part of a location name is mentioned? (Section~\ref{sec:LocPart})
    \item Is there an example of disjointed organization names where the mentions are self-sufficient? (Section~\ref{sec:OrgDisjoint})
\end{itemize}

Some questions still open:
\begin{itemize}
    \item How to annotate adjectives derived from the proper noun of persons? e.g. Marxist, Circean, etc. % ex. D'Artagnan was again in the presence of the Circean who had before surrounded him with her enchantments. (The Three Musketeers)
\end{itemize}

%%%%%%%%%%%%%%%%%%%%%%%%%%%%%%%%%%%%%%%%%%%%%%%%%%%%
\newpage
\phantomsection\addcontentsline{toc}{section}{References}

%%%%%%%%%%%%%%%%%%%%%%%%%%
% biblatex (much slower, for some reason)
% \printbibliography

%%%%%%%%%%%%%%%%%%%%%%%%%%
% bibtex (natbib)
\footnotesize
\setlength{\bibsep}{3pt plus 1.5ex}
\bibliography{novelties_biblio.bib}
%%%% natbib doc:
% https://texlive.mycozy.space/macros/latex/contrib/natbib/natnotes.pdf
% citep >> [1]
% citet >> Author [1]
% \citeauthor >> Author 
% \citeyear >> 1990

\end{document}